%% file: main.tex
\pdfoutput=1
\documentclass[11pt]{article}
\usepackage{ACL2023}

\usepackage{times}
\usepackage{latexsym}
\usepackage[T2A,T1]{fontenc}
\usepackage[russian,english]{babel}
\usepackage[utf8]{inputenc}
\usepackage{microtype}
\usepackage{inconsolata}
\usepackage{graphicx}
\usepackage{multirow}
\usepackage{multicol}
\usepackage{colortbl}
\usepackage{array}
\usepackage{mathtools}
\usepackage{rotating}
\usepackage{tabularray}
\usepackage{vcell}
\title{Contextual Cues in Machine Translation: Investigating the Potential of Multi-Source Input Strategies in LLMs and NMT Systems}

\author{Lia Shahnazaryan \and Patrick Simianer \and Joern Wuebker \\
LILT \\
\texttt{\{lia,patrick,joern\}@lilt.com}\\}

\begin{document}
\maketitle
\begin{abstract}
We explore the impact of multi-source input strategies on machine translation (MT) quality, comparing GPT-4o, a large language model (LLM), with a traditional multilingual neural machine translation (NMT) system. Using intermediate language translations as contextual cues, we evaluate their effectiveness in enhancing English and Chinese translations into Portuguese. Results suggest that contextual information significantly improves translation quality for domain-specific datasets and potentially for linguistically distant language pairs, with diminishing returns observed in benchmarks with high linguistic variability. Additionally, we demonstrate that shallow fusion, a multi-source approach we apply within the NMT system, shows improved results when using high-resource languages as context for other translation pairs, highlighting the importance of strategic context language selection.
\end{abstract}

\section{Introduction}
\label{sec:introduction}
Machine translation (MT) continues to evolve with advances in neural network architectures and large language models (LLMs) \citep{kocmi-etal-2024-findings}. Traditional neural machine translation (NMT) systems have explored strategies such as multi-source and multi-way MT to improve translation quality \citep{zoph-knight-2016-multi, firat-etal-2016-multi}, while LLMs leverage in-context learning (ICL) \citep{dong2022survey} and various prompting strategies \citep{zhang2023prompting}. These developments open up new opportunities for improving translation workflows, particularly by leveraging translations into intermediate languages as additional context to enhance the quality of subsequent translations in multilingual processing pipelines. 

This research examines the potential of incorporating multi-source input to improve MT quality, comparing LLMs and traditional NMT systems. While this approach has been explored in the past with traditional NMT systems, we extend it to LLMs. Specifically, the work investigates how intermediate language translations can be used as contextual information to enhance subsequent translations. The study addresses two core research questions:
\begin{enumerate}
    \item Can multi-source input be effectively leveraged to enhance MT quality? 
    \item How do LLMs, such as GPT-4o, compare to multilingual NMT systems in utilizing multi-source input for enhancing translation performance? 
\end{enumerate}
To explore these questions, we conduct a series of experiments using both an LLM and a custom-built multilingual NMT system. Our approach evaluates direct source-target translations, the use of single and multiple context languages both with human-edited gold standard and LLM-generated translations, and shallow fusion to incorporate multi-source input into the NMT system. The study focuses on English and Chinese as source languages, Portuguese as the target language, and Spanish, French, Italian, German, and Russian as context languages. Evaluations are performed on proprietary technical datasets, as well as established benchmarks, providing insights into domain-specific and general translation performance. The key contributions of this work include: 
\begin{itemize}
    \item Application of a shallow fusion method within a single multilingual NMT system, showing significant improvements when an optimal intermediate language is used across source-target language pairs.
    \item A detailed comparative evaluation of GPT-4o and a custom NMT system, providing insights into their respective strengths and limitations in terms of adaptability in leveraging contextual information across diverse scenarios.
    \item Identification of conditions where contextual information improves translation quality, particularly in domain-specific datasets, while also recognizing potential drawbacks in benchmarks with diverse linguistic variability. 
\end{itemize}

\section{Related Work}
\label{sec:related_work}
The concept of multi-source MT, which leverages information from multiple source languages, has been an active area of research in NMT for several years. Early research demonstrates the potential of combining information from diverse source languages to improve target language translations \citep{garmash-monz-2016-ensemble, dabre-etal-2017-enabling, libovicky-etal-2018-input}. Shallow fusion, on the other hand, initially proposed by \cite{gulcehre2015using}, typically involves using an external target language model during decoding through a weighted log-linear combination of the translation and language model output probabilities \citep{subramanian-etal-2021-nvidia}. Our approach diverges from this by integrating multi-source input directly within a single multilingual many-to-many NMT model. Rather than using an external model, we combine log probabilities from multiple source languages during decoding to select the best translation hypothesis. 

Recent advances in LLMs have also significantly impacted MT, with models demonstrating strong zero-shot and few-shot translation capabilities \citep{hendy2023good, zhu-etal-2024-multilingual}. Building on these advances, researchers have explored various prompting strategies to optimize LLM performance in translation tasks \citep{vilar-etal-2023-prompting, zhang-etal-2023-machine}. ICL, a key capability of LLMs, has also been applied to MT research. For example, \citet{zhu-etal-2024-multilingual} demonstrate that incorporating cross-lingual exemplars in prompts has the potential to improve translation quality. Our study builds upon this body of work by taking a different approach: prompting the model with the source sentence and its translations in context languages to assess how contextual information impacts translation accuracy.

\section{Setup}
\label{sec:setup}
Our experimental setup is designed to systematically evaluate the effectiveness of leveraging multi-source input for improving MT quality. In this section, we compare various translation approaches by also outlining the datasets, models, and evaluation metrics employed in the study. The experiments are structured to assess the impact of single and multiple context languages, as well as the impact of sequential translation and shallow fusion approaches.

\subsection{Datasets}
\label{sec:datasets}
For our experiments, we use five datasets: three proprietary datasets from internal technical domains and two evaluation benchmarks. In all cases, English serves as the original source language. This selection allows us to assess the performance of our translation strategies across various domains.

\subsubsection{Proprietary Datasets}
We use three proprietary datasets from internal sources, each representing technical domains. These datasets contain 3,000 test sentences and provide translations in all context languages used in our experiments, including Spanish, French, Italian, German, and Russian, as well as the English and Chinese sources, and the Portuguese target.

\subsubsection{Evaluation Benchmarks}
\textbf{FLORES+}: The evaluation benchmark \citep{costa2022no} is a multilingual dataset covering 200 languages. It is designed to evaluate MT quality across a wide range of languages and domains. The test set (devtest) includes 1,012 sentences, providing translations in all source, context, and target languages used in our experiments.
\newline
\textbf{TICO-19}: The evaluation benchmark \citep{anastasopoulos-etal-2020-tico} focuses on the medical domain, specifically COVID-19-related content. It includes 2,100 test sentences. While it does not cover all our context languages, it does provide translations in Spanish, French, and Russian.

\subsection{Models}
\label{sec:models}
Our study employs two types of models: an LLM and a multilingual NMT system.

\subsubsection{GPT-4 LLM}
For the LLM-based translations, we use OpenAI's GPT-4 model \citep{achiam2023gpt}, specifically the GPT-4 omni (GPT-4o) variant \citep{gpt4o}, which is a decoder-only Transformer model \citep{vaswani2017attention}, optimized for language generation tasks and requiring no task-specific fine-tuning. We utilize the model's few-shot translation abilities, prompting it with source text and context from other languages to generate translations. We use a series of structured prompts, adaptable to varying numbers of context languages. Table \ref{fig:prompt-templates} illustrates basic structure of these prompts. 
\begin{table*}[ht]
\centering
\resizebox{\textwidth}{!}{%
\begin{tabular}{l}
\textbf{Prompt 1: Direct Source-to-Target Translation} \\ \hline
\begin{tabular}[c]{@{}l@{}}Translate from \{source\_language\} to \{target\_language\}.\\
Output only the translated sentence.\\ \{source\_language\} SOURCE: \{source\_sentence\}\\ \{target\_language\} TRANSLATION:\end{tabular} \\
\\
\textbf{Prompt 2: Translation with Single Context Language} \\ \hline
\begin{tabular}[c]{@{}l@{}}Translate from \{source\_language\} to \{target\_language\}, given the translation in \{context\_language\}. \\
Output only the translated sentence.\\ \{source\_language\} SOURCE: \{source\_sentence\}\\ \{context\_language\} CONTEXT: \{context\_sentence\}\\ \{target\_language\} TRANSLATION:\end{tabular} \\
\\
\textbf{Prompt 3: Translation with Multiple Context Languages} \\ \hline
\begin{tabular}[c]{@{}l@{}}Translate from \{source\_language\} to \{target\_language\}, given the translations in \{context\_language\_1\} and \{context\_language\_2\}. \\
Output only the translated sentence.\\ \{source\_language\} SOURCE: \{source\_sentence\}\\ \{context\_language\_1\} CONTEXT 1: \{context\_sentence\_1\}\\ \{context\_language\_2\} CONTEXT 2: \{context\_sentence\_2\}\\ \{target\_language\} TRANSLATION:\end{tabular}
\end{tabular}%
}
\caption{Prompt templates used for GPT-4o translation tasks with varying numbers of context languages. This pattern is extended for experiments involving three context languages.}
\label{fig:prompt-templates}
\end{table*}

\subsubsection{Multilingual NMT System}
\label{sec:nmt}
The NMT system employed in the experiments is a commercially used model, built using the NVIDIA NeMo toolkit \citep{kuchaiev2019nemo}. Fundamentally, the model is a traditional encoder-decoder Transformer \citep{vaswani2017attention} with 21 encoder- and two decoder layers, overall comprising about 1.3B parameters. Being a multilingual system, the model supports translation between all combinations of the following languages: English, German, French, Italian, Spanish, and Portuguese. The data used to train the model consists of a wide range of publicly available and private datasets, but it does not include any of the datasets used for validation and testing in the experiments presented here.

We also implement shallow fusion, by integrating the primary source input \( x \) and one or more optional context inputs \( z_1 \), \( z_2 \), ..., \( z_n \) during decoding. At each decoding step, the model generates log probabilities based on both the primary source and context inputs. The final score for each translation hypothesis is computed as:
\begin{align}
\text{score} = \lambda_0 \log P(y \mid x) + \sum_{i=1}^{n} \lambda_i \log P(y \mid z_i)
\end{align}
where \( y \) is the target sequence, \( \lambda_0 \) and \( \lambda_i \) are fusion coefficients controlling the influence of the source input and each context input, respectively, and \( n \) is the number of context languages. In our experiments, \( \lambda_i = 1, \forall i \in \{0, 1,...,n\} \), giving equal weights to the source and context inputs (preliminary experiments with different weights yielded scores within the performance bounds established by the source-target and context-target baselines, thus we report results only for equal weights).

\subsection{Experiments}
\label{sec:experiments}
We conduct experiments focusing on three key areas, each designed to evaluate specific aspects of our translation strategies and their effectiveness in various contexts.

\subsubsection{Direct vs. Contextual Translations with GPT-4o}
In this experiment, we aim to assess the impact of contextual information on GPT-4o's translation quality. We establish a baseline using direct source-to-target translations without additional context. We then incrementally introduce contextual information in two phases:
\begin{itemize}
    \item \textbf{Single context language}: We provide GPT-4o with the source sentence and its translation in one context language (Spanish, French, Italian, German, or Russian).
    \item \textbf{Multiple context languages}: We extend the input to include translations in multiple languages (e.g., Spanish and French, or Spanish, French, and Italian).
\end{itemize}
By comparing these approaches to the baseline, we aim to evaluate the overall influence of contextual information on GPT-4o's translation performance, the relative effectiveness of single vs. multiple context languages, and the potential variations in performance across different language combinations.

\subsubsection{Sequential Translation Experiments}
We extend the study of contextual information's impact on GPT-4o's translation quality by simulating real-world scenarios where context translations are generated by the LLM itself. We compare this sequential approach to both the baseline and the contextual translation experiments described previously. First, we use GPT-4o to translate the source sentence into a context language (e.g., Spanish). This model-generated translation, without any post-editing, is then used as context when translating into the target language.

By comparing this sequential approach to both the baseline and the previous contextual translation experiments, we aim to evaluate how the quality of the LLM-generated intermediate translation affects the final translation compared to using gold standard context translations. We also assess the model's ability to use its own generated content as context for subsequent translations, and how this compares to its performance with externally provided context. This provides insights into the practical applicability of using LLMs in multi-step translation processes, simulating scenarios where human-edited translations may not be available as context.

\subsubsection{Comparison between GPT-4o and Traditional NMT System}
To provide a comprehensive evaluation of our approach, we also conduct a comparative study between GPT-4o and an NMT system. This experiment provides insights into the potential of LLMs in translation tasks and helps us understand how they compare to established NMT systems in terms of translation quality, adaptability to context, and overall performance.

We start with direct source-to-target translations using the NMT system, which follows the same methodology as the baseline established for GPT-4o. This initial step allows us to quantify performance differences between the two approaches under standard translation conditions. We then implement shallow fusion within the NMT system, as described in Section \ref{sec:nmt}. This step mirrors our contextual experiments with GPT-4o. By integrating contextual information into the NMT model, we assess whether this hybrid approach yields improvements in translation quality over the direct NMT results.

\subsection{Metrics}
\label{sec:metrics}
We use BLEU \citep{papineni-etal-2002-bleu}, implemented through the SacreBLEU framework \citep{post-2018-call} with the default 13a tokenizer, to evaluate translation quality across the experiments. We also employ COMET \citep{rei-etal-2020-comet}, a reference-based metric using neural models, and its reference-free variant COMETKIWI \citep{rei-etal-2022-cometkiwi} to capture semantic accuracy. To ensure the reliability of the comparisons, we conduct statistical significance testing using sacreBLEU's and COMET's implementation of paired bootstrap resampling \citep{koehn-2004-statistical}, with a significance threshold of 0.05.

\section{Results and Discussion}
\label{sec:results_discussion}
\subsection{Experimental Results}
The results are presented in Tables \ref{tab:results_1} and \ref{tab:results_2}, covering the English-to-Portuguese and Chinese-to-Portuguese translation tasks across the proprietary, TICO-19, and FLORES+ datasets. Results for the proprietary datasets are averaged, as they all represent the same domain (see Appendix \ref{appd:full_results} for detailed results across individual datasets). 
\begin{table*}[ht]
\centering
\resizebox{\textwidth}{!}{%
\begin{tabular}{l|l|lll|lll|lll}
 &  & \multicolumn{3}{c|}{\textbf{PROPRIETARY (AVG)}} & \multicolumn{3}{c|}{\textbf{TICO-19}} & \multicolumn{3}{c}{\textbf{FLORES+}} \\
Model/Experiment & CONTEXT & BLEU & COMET & COMETKIWI & BLEU & COMET & COMETKIWI & BLEU & COMET & COMETKIWI \\ 
\hline
GPT-4o Baseline & NONE & 50.12 & 89.72 & 81.13 & \textbf{53.01} & \textbf{90.36} & 85.50 & \textbf{51.53} & \textbf{90.59} & 85.73 \\ 
\hline
\multirow{8}{*}{GPT-4o Contextual} & ES & 45.86 & 88.20 & 79.57 & 47.86 & 89.45 & 85.45 & 34.53 & 88.64 & 85.38 \\
 & FR & 48.35 & 89.65 & 80.77 & 43.70 & 85.17 & 80.60 & 43.62 & 89.71 & 85.81 \\
 & IT & 47.96 & 90.07 & 80.52 & - & - & - & 38.48 & 89.11 & 85.43 \\
 & DE & 49.12 & 89.83 & 80.78 & - & - & - & 48.32 & 90.27 & 85.78 \\
 & RU & 46.94 & 89.47 & 80.46 & 47.48 & 89.34 & 85.14 & 46.65 & 89.97 & 85.58 \\
 & ES + FR & 51.57 & 90.50 & 81.31 & 51.73 & 90.04 & 85.55 & 48.59 & 90.50 & \textbf{86.03} \\
 & FR + IT & 51.39 & 90.53 & 81.10 & - & - & - & 48.85 & 90.46 & 85.88 \\
 & ES + FR + IT & \textbf{52.43} & \textbf{90.73} & 81.31 & - & - & - & 49.14 & 90.57 & 86.02 \\ 
\hline
\multirow{8}{*}{GPT-4o Sequential} & ES & 48.18 & 89.39 & 81.33 & 50.87 & 90.07 & 85.55 & 48.81 & 90.23 & 85.81 \\
 & FR & 48.06 & 89.40 & 81.29 & 50.98 & 90.16 & \textbf{85.62} & 48.02 & 90.16 & 85.81 \\
 & IT & 48.47 & 89.60 & 81.06 & - & - & - & 48.00 & 90.12 & 85.70 \\
 & DE & 48.85 & 89.63 & 81.30 & - & - & - & 49.72 & 90.34 & 85.76 \\
 & RU & 48.22 & 89.53 & 81.20 & 50.64 & 90.06 & 85.55 & 49.21 & 90.38 & 85.81 \\
 & ES + FR & 49.60 & 89.59 & \textbf{81.43} & 52.34 & 90.31 & \textbf{85.62} & 49.98 & 90.44 & 85.90 \\
 & FR + IT & 49.56 & 89.73 & 81.35 & - & - & - & 49.74 & 90.41 & 85.80 \\
 & ES + FR + IT & 50.01 & 89.75 & 81.35 & - & - & - & 50.26 & 90.51 & 85.92 \\
 \hline
 \hline
NMT Baseline & NONE & \textbf{53.85} & \textbf{90.14} & \textbf{81.55} & \textbf{51.79} & \textbf{90.07} & \textbf{85.62} & \textbf{53.02} & \textbf{90.52} & \textbf{85.91} \\ 
\hline
\multirow{5}{*}{NMT Shallow Fusion} & ES & 51.61 & 89.71 & 81.21 & 48.91 & 89.53 & 85.52 & 45.82 & 89.84 & 85.34 \\
 & FR & 51.40 & 89.69 & 81.11 & 43.46 & 85.75 & 80.50 & 48.91 & 90.05 & 85.34 \\
 & IT & 50.95 & 89.72 & 80.99 & - & - & - & 47.00 & 89.87 & 85.46 \\
 & DE & 50.44 & 89.54 & 81.13 & - & - & - & 48.00 & 90.10 & 85.53 \\
 & ES + FR + IT & 51.27 & 89.93 & 81.24 & - & - & - & 44.83 & 89.76 & 85.51
\end{tabular}%
}
\caption{English-to-Portuguese translation results using GPT-4o and an NMT system, evaluated on proprietary, TICO-19, and FLORES+ datasets using BLEU, COMET, and COMETKIWI. GPT-4o is tested in baseline, single and multiple context language setups, and a sequential setup with intermediate context generation. The NMT baseline uses direct translation, while shallow fusion combines predictions from multiple sources during decoding (see Section \ref{sec:nmt}). Proprietary results are averaged across three datasets (see Appendix \ref{appd:full_results} for detailed scores).}
\label{tab:results_1}
\end{table*}
\newline
\newline
\textbf{Direct vs. Contextual vs. Sequential Translations with GPT-4o}:
In the English-to-Portuguese translation task, the baseline results using GPT-4o without context information demonstrate strong performance, achieving the highest BLEU and COMET scores across two benchmarks: FLORES+ and TICO-19 (Table \ref{tab:results_1}). Introducing contextual information from single or multiple languages leads to lower performance in both cases. The only exception is the COMETKIWI results for FLORES+, where the inclusion of two or more context languages leads to better performance. For TICO-19, however, the gains in COMETKIWI scores are not significant. On proprietary datasets, the inclusion of context from multiple languages (specifically, different combinations of Spanish, French, and Italian) yields improvements over the baseline, highlighting the potential benefits of multi-source context in domain-specific scenarios.

For the sequential translation approach, where GPT-4o generates its own context translations, the results are less promising in terms of BLEU and COMET (see Table \ref{tab:results_1}). While this approach demonstrates marginal improvements over the baseline for certain proprietary datasets, the general trend indicates a decline in performance. However, the approach shows greater stability across different context languages when only a single context language is used. For example, experiments using Spanish or Russian as single context languages result in notable improvements compared to the same experiments within the contextual approach (48.18 BLEU in sequential vs. 45.86 BLEU in contextual for Spanish, and 48.22 BLEU in sequential vs. 46.94 BLEU in contextual for Russian on proprietary datasets). In contrast, when multiple context languages are combined, the sequential translation approach does not replicate the performance gains observed with the contextual approach in proprietary datasets. For the FLORES+ and TICO-19 datasets, however, this approach shows a slightly different pattern. While its performance remains below the baseline, it generally outperforms the contextual approach. 
\begin{table*}[ht]
\centering
\resizebox{\textwidth}{!}{%
\begin{tabular}{l|l|lll|lll|lll}
 &  & \multicolumn{3}{c|}{\textbf{PROPRIETARY (AVG)}} & \multicolumn{3}{c|}{\textbf{TICO-19}} & \multicolumn{3}{c}{\textbf{FLORES+}} \\
Model/Experiment & CONTEXT & BLEU & COMET & COMETKIWI & BLEU & COMET & COMETKIWI & BLEU & COMET & COMETKIWI \\ 
\hline
GPT-4o Baseline & NONE & 30.37 & 86.25 & \textbf{78.51} & 29.54 & 87.20 & \textbf{83.21} & 26.48 & 87.60 & \textbf{83.64} \\ 
\hline
\multirow{2}{*}{GPT-4o Contextual} & ES & 43.34 & 89.18 & 75.28 & 44.38 & 89.14 & 80.39 & 27.85 & 87.99 & 79.69 \\
 & ES + FR & \textbf{46.05} & \textbf{89.73} & 76.10 & \textbf{44.76} & \textbf{89.48} & 81.36 & \textbf{37.62} & \textbf{89.45} & 81.80
\end{tabular}%
}
\caption{Results for GPT-4o in various contextual setups for Chinese-to-Portuguese translation, evaluated on proprietary, TICO-19, and FLORES+ datasets using BLEU, COMET, and COMETKIWI. The proprietary results are averaged across three datasets (see Appendix \ref{appd:full_results} for detailed scores). Experiments include direct translation without context (baseline), as well as with single and multiple context languages.}
\label{tab:results_2}
\end{table*}

In addition to the English-to-Portuguese contextual experiments, we further extend the evaluation to the Chinese-to-Portuguese translation task. As shown in Table \ref{tab:results_2}, adding context results in significant improvements across all datasets, as reflected in both BLEU and COMET scores. The performance further improves with the inclusion of additional context languages. This demonstrates that multi-source input may be particularly beneficial for linguistically distant language pairs, though further investigation is needed to confirm this across a broader range of languages. However, COMETKIWI scores consistently decline with the addition of contextual information, revealing an important aspect of reference-free evaluation: its sensitivity to source language adherence. 
\begin{table*}[ht]
\centering
\resizebox{\textwidth}{!}{%
\begin{tabular}{l|l|lll|lll|lll}
\vcell{} & \vcell{} & \multicolumn{3}{c|}{\vcell{\textbf{PROPRIETARY (AVG)}}} & \multicolumn{3}{c|}{\vcell{\textbf{TICO-19}}} & \multicolumn{3}{c}{\vcell{\textbf{FLORES+ }}} \\[-\rowheight]
\printcellbottom & \printcellbottom & \multicolumn{3}{c|}{\printcellmiddle} & \multicolumn{3}{c|}{\printcellmiddle} & \multicolumn{3}{c}{\printcellmiddle} \\
Model \textbar{} Source & CONTEXT & BLEU & COMET & COMETKIWI & BLEU & COMET & COMETKIWI & BLEU & COMET & COMETKIWI \\ 
\hline
Baseline \textbar{} EN & NONE & \textbf{53.85} & \textbf{90.14} & \textbf{81.55} & \textbf{51.79} & \textbf{90.07} & \textbf{85.62} & \textbf{53.02} & \textbf{90.52} & \textbf{85.91} \\ 
\hline\hline
Baseline \textbar{} ES & NONE & 46.49 & 89.74 & \textbf{78.89} & 45.34 & 89.59 & \textbf{84.69} & 28.73 & 88.34 & \textbf{85.19} \\
Shallow Fusion \textbar{} ES & EN & \textbf{51.61} & \textbf{90.09} & 77.78 & \textbf{48.91} & \textbf{89.85} & 84.03 & \textbf{45.82} & \textbf{89.93} & 83.28 \\ 
\hline
Baseline \textbar{} FR & NONE & 45.98 & 88.73 & \textbf{79.35} & 36.79 & 83.01 & \textbf{85.02} & 39.68 & 88.94 & \textbf{85.08} \\
Shallow Fusion \textbar{} FR & EN & \textbf{51.40} & \textbf{89.55} & 78.13 & \textbf{43.46} & \textbf{85.19} & 78.02 & \textbf{48.91} & \textbf{89.84} & 83.97 \\ 
\hline
Baseline \textbar{} IT & NONE & 45.41 & 89.18 & \textbf{80.13} & - & - & - & 32.29 & 88.50 & \textbf{85.67} \\
Shallow Fusion \textbar{} IT & EN & \textbf{50.95} & \textbf{89.85} & 78.87 & - & - & - & \textbf{47.00} & \textbf{89.94} & 83.94 \\ 
\hline
Baseline \textbar{} DE & NONE & 42.34 & 87.70 & \textbf{80.34} & - & - & - & 37.32 & 88.05 & \textbf{85.25} \\
Shallow Fusion \textbar{} DE & EN & \textbf{50.44} & \textbf{88.98} & 78.99 & - & - & - & \textbf{48.00} & \textbf{89.31} & 84.45
\end{tabular}%
}
\caption{Comparison of baseline NMT systems and shallow fusion setups across different source languages translating into Portuguese on proprietary (averaged), TICO-19, and FLORES+ datasets, using BLEU, COMET, and COMETKIWI.}
\label{tab:nmt_detailed}
\end{table*}
\newline
\newline
\textbf{Comparison between GPT-4o and Traditional NMT System}: The NMT system baseline generally outperforms GPT-4o in direct source-to-target translations for the English-to-Portuguese task across most datasets and metrics (Table \ref{tab:results_1}). The only exception is the TICO-19 dataset, where GPT-4o surpasses the NMT system baseline by 1.22 BLEU and 0.29 COMET. However, on the proprietary datasets, GPT-4o with contextual information outperforms the NMT baseline when evaluated using COMET. Specifically, the setup with three context languages (Spanish, French, Italian) achieves an average COMET improvement of 0.59 over the NMT system. Similarly, on the FLORES+ dataset, using Spanish and French as context languages results in a 0.12 gain over the NMT baseline in COMETKIWI evaluations.

In contrast, incorporating shallow fusion into the NMT system to leverage contextual information for the English-to-Portuguese translation task does not yield improvements, as seen in Table \ref{tab:results_1}. The results fall short of the NMT baseline, suggesting that the NMT model is already optimized for translations with English as the source language, thus deriving little benefit from additional context.

However, shallow fusion also reveals an important performance pattern when applied to other language directions. Table \ref{tab:nmt_detailed} shows that shallow fusion improves translation performance for all source languages (other than English) within the same multilingual model when evaluated using BLEU and COMET. Specifically, it enhances performance when English is used as context for translating non-English source languages into Portuguese, while it leads to a decline in performance for English-to-Portuguese translations (see Table \ref{tab:results_1}). This suggests that using English context to improve translations from other source languages into Portuguese boosts performance, demonstrating that multilingual models can be tailored to leverage the language they best understand to improve translations into other available language pairs.

\subsection{Impact of Context on GPT-4o Translations}
Looking at Tables \ref{tab:results_1} and \ref{tab:results_2}, we observe distinct patterns in how contextual information affects translation quality across language pairs and datasets. 
\begin{table*}[ht]
\centering
\resizebox{\textwidth}{!}{%
\begin{tabular}{llll}
\multicolumn{1}{l|}{\textbf{Source}} & \multicolumn{1}{l|}{\textbf{Context}} & \multicolumn{1}{l|}{\textbf{Translation}} & \textbf{Reference} \\ \hline
\multicolumn{4}{l}{\textbf{FLORES+ (General Domain)}} \\ \hline
\multicolumn{1}{l|}{\multirow{2}{*}{cases of the Ebola virus}} & \multicolumn{1}{l|}{N/A} & \multicolumn{1}{l|}{casos do vírus Ebola} & \multirow{2}{*}{casos do vírus Ebola} \\ \cline{2-3}
\multicolumn{1}{c|}{} & \multicolumn{1}{l|}{casos de Ébola (Spanish)} & \multicolumn{1}{l|}{casos de Ebola} &  \\
\multicolumn{4}{l}{} \\
\textbf{Proprietary (Technical Domain)\textsuperscript{†}} \\ \hline
\multicolumn{1}{l|}{\multirow{2}{*}{Vortex Chain}} & \multicolumn{1}{l|}{N/A} & \multicolumn{1}{l|}{Cadeia Vortex} & \multirow{2}{*}{Vortex Chain} \\ \cline{2-3}
\multicolumn{1}{l|}{} & \multicolumn{1}{l|}{Vortex Chain (Russian)} & \multicolumn{1}{l|}{Vortex Chain} & 
\end{tabular}%
}
\caption{Impact of context on terminology consistency. In proprietary domains, context helps maintain consistent terminology, while in general text (FLORES+), it may introduce stylistic variations common in different languages. \textsuperscript{†} indicates that the proprietary example shown here is synthetic but representative of the patterns observed in the actual proprietary datasets.}
\label{tab:terminology_example}
\end{table*}
For English-to-Portuguese translation (Table \ref{tab:results_1}), the benefits of contextual information are evident primarily in proprietary datasets. Here, the addition of multiple context languages leads to improvements of 2.31 BLEU, 1.01 COMET, and 0.18 COMETKIWI. In contrast, for the FLORES+ and TICO-19 benchmarks, direct translation without context appears to yield superior results, with the only exception being slight improvements over the baseline observed in COMETKIWI scores.

This disparity can be attributed to the nature of the datasets: the proprietary datasets contain domain-specific technical content with consistent terminology and structures, which likely benefit from seeing translations of the same terms across different context languages, while the more general (FLORES+) and medical (TICO-19) datasets may show greater variability in how sentences are translated across languages, with less consistency in terminology (see Table \ref{tab:terminology_example}). As a result, the additional context might introduce noise rather than providing helpful information, particularly in reference-based evaluations. Furthermore, there is a clear trend in how the number of context languages impacts translation quality.

While using a single context language, irrespective of which language, often results in mixed outcomes and generally underperforms compared to the baseline (with a few exceptions when evaluated using COMET), using multiple context languages consistently improves performance over both the baseline and single-context scenarios for proprietary datasets. This suggests that having multiple contextual sources offers richer linguistic cues, which can be particularly beneficial for disambiguating terms in specialized domains by providing additional information on terminology and phraseology.

The sequential approach reveals distinct trends, particularly regarding translation consistency across different context languages. Table \ref{tab:examples} highlights a key factor influencing this consistency: using a single context language in the sequential approach results in more stable outcomes regardless of the language, compared to the contextual approach, which shows greater variability depending on the context language. For instance, BLEU scores for single-context languages in the sequential approach are closely aligned (e.g., 48.81 for Spanish, 48.02 for French, and 49.21 for Russian on FLORES+). However, the contextual approach shows wider variability for the same context languages and dataset (e.g., 34.53 for Spanish, 43.62 for French, and 46.65 for Russian). Table \ref{tab:examples} provides examples illustrating these differences, highlighting how human-generated context translations in the contextual approach contribute to this instability in scores across experiments. 

As shown, human-generated context translations introduce stylistic variability, often causing target translations to diverge from the reference, which tends to be syntactically closer to the source. This explains why some context languages result in significantly lower scores in the contextual approach, whereas the sequential approach produces more uniform scores. Furthermore, adding multiple LLM-generated context translations in the sequential setup does not improve performance. This suggests that human-generated context translations offer more nuanced contextual information and greater variability, which LLM-generated translations lack. Consequently, the sequential approach yields consistent scores regardless of the number or choice of context languages added.

\begin{table*}[ht]
\centering
\resizebox{\textwidth}{!}{%
\begin{tabular}{p{0.18\textwidth}|p{0.45\textwidth}|p{0.12\textwidth}|p{0.45\textwidth}}
\multicolumn{4}{l}{\textbf{Russian Context Example }} \\ 
\hline
\textbf{Source} & A curry can be either "dry" or "wet" depending on the amount of liquid. & \textbf{Reference} & Um curry pode ser "seco" ou "molhado" dependendo da quantidade de líquido. \\
\hline
\textbf{Context (Human)} & \foreignlanguage{russian}{\colorbox[rgb]{0.961,0.8,0.8}{В зависимости от содержания жидкости}, карри может быть «сухим» или «мокрым».} & \textbf{Translation} & \colorbox[rgb]{0.961,0.8,0.8}{Dependendo do conteúdo de líquido}, o curry pode ser "seco" ou "molhado". \\
\hline
\textbf{Context (LLM)} & \foreignlanguage{russian}{Карри может быть либо «сухим», либо «жидким» в зависимости от количества жидкости.} & \textbf{Translation} & Um curry pode ser "seco" ou "molhado" dependendo da quantidade de líquido. \\
\multicolumn{4}{l}{} \\
\multicolumn{4}{l}{\textbf{Spanish Context Example }} \\ 
\hline
\textbf{Source} & All citizens of Vatican City are Roman Catholic. & \textbf{Reference} & Todos os cidadãos da cidade do Vaticano são católicos romanos. \\
\hline
\textbf{Context (Human)} & \colorbox[rgb]{0.961,0.8,0.8}{La totalidad de los ciudadanos que viven en} Ciudad del Vaticano \colorbox[rgb]{0.961,0.8,0.8}{adscriben a la religión católica romana}. & \textbf{Translation} & \colorbox[rgb]{0.961,0.8,0.8}{A totalidade dos cidadãos que vivem na} Cidade do Vaticano \colorbox[rgb]{0.961,0.8,0.8}{adere à religião católica romana}. \\
\hline
\textbf{Context (LLM)} & Todos los ciudadanos de la Ciudad del Vaticano son católicos romanos. & \textbf{Translation} & Todos os cidadãos da Cidade do Vaticano são católicos romanos.
\end{tabular}%
}
\caption{Examples of English-to-Portuguese translations from the FLORES+ benchmark, using Russian and Spanish context generated by humans vs. GPT-4o. The translations are compared to the reference within contextual and sequential approaches. Highlighted sections show where human-generated context (contextual approach) deviates from the English source and LLM-generated context (sequential approach), illustrating how these variations impact translation accuracy relative to the Portuguese reference.}
\label{tab:examples}
\end{table*}
The impact of contextual information, however, is more pronounced in Chinese-to-Portuguese translation (Table \ref{tab:results_2}). Here, context provides consistent improvements across all datasets, suggesting that contextual information may play a greater role when translating between linguistically more distant language pairs. A possible explanation is that additional context helps mitigate structural and lexical differences between Chinese and Portuguese, leading to more accurate translations. However, given the limited scope of our evaluation, further research is needed to determine whether this trend holds across other typologically distant language pairs.

In contrast, the progressive decrease in COMETKIWI scores with added context implies that it penalizes translations that deviate from source semantics, even when these deviations actually improve the naturalness of the target language output. While intermediate languages help bridge the linguistic gap (as seen in BLEU and COMET improvements), they may lead to translations that prioritize target language conventions. This highlights a trade-off between source fidelity and target language fluency.

\section{Conclusion}
\label{sec:conclusion}
This study analyzes how multi-source input strategies influence MT performance, comparing GPT-4o with a custom-trained multilingual NMT system. We show that contextual cues from multiple intermediate languages significantly enhance translation quality for technical domains with defined terminology. The experiments also suggest this approach may be particularly beneficial for linguistically distant language pairs, though further research with additional language pairs would be needed to validate this hypothesis. When GPT-4o generates its own intermediate translations for context, performance remains consistently below baseline levels, suggesting that the model does not provide additional language-specific information comparable to gold-standard context translations. The comparison between GPT-4o and the NMT system highlights their complementary strengths: while the latter excels in direct English-to-Portuguese translations, particularly on proprietary datasets, GPT-4o shows superior performance in leveraging contextual information for domain-specific content. Furthermore, our implementation of shallow fusion in the NMT system enhances the model's performance for non-English source languages by effectively leveraging English as an auxiliary context, suggesting that multilingual models can be optimized by leveraging their strongest language pair to enhance performance across other language combinations. 

These findings demonstrate the potential of multi-source strategies to enhance translation accuracy across diverse scenarios. They emphasize the importance of selecting models and methods based on task-specific requirements, such as leveraging GPT-4o for its contextual adaptability and using shallow fusion in multilingual NMT systems for non-English source-target pairs. 

\section{Limitations}
\label{sec:limitations}
Our study has several limitations. It focuses on a limited set of languages, with English and Chinese as source languages and Portuguese as the target. This narrow scope may not fully represent the potential benefits or challenges of multi-source approaches across more diverse language families and typological relationships. Additionally, evaluating the computational costs or latency implications of generating and combining multiple translations in production environments, particularly in LLM-based systems, is beyond the scope of this study. Future research should address these aspects by expanding language coverage and assessing multi-source integration in real-world applications.

\bibliography{anthology,custom}
\bibliographystyle{acl_natbib}
\appendix
\section{Complete Evaluation Results}
\label{appd:full_results}
\input{appendix}

\end{document}

%% file: appendix.tex
\begin{sidewaystable*}
\centering
\resizebox{\textwidth}{!}{%
\begin{tabular}{l|l|lll|lll|lll|lll|lll}
 &  & \multicolumn{3}{c|}{\textbf{FLORES+}} & \multicolumn{3}{c|}{\textbf{TICO-19}} & \multicolumn{3}{c|}{\textbf{PROPRIETARY A}} & \multicolumn{3}{c|}{\textbf{PROPRIETARY B}} & \multicolumn{3}{c}{\textbf{PROPRIETARY C}} \\
Model/Experiment & CONTEXT & BLEU & COMET & COMETKIWI & BLEU & COMET & COMETKIWI & BLEU & COMET & COMETKIWI & BLEU & COMET & COMETKIWI & BLEU & COMET & COMETKIWI \\ 
\hline
GPT-4o Baseline & NONE & \textbf{51.53} & \textbf{90.59} & 85.73 & \textbf{53.01} & \textbf{90.36} & 85.50 & 51.49 & 91.04 & 81.40 & 48.22 & 88.08 & 77.67 & 50.65 & 90.04 & 84.32 \\ 
\hline
\multirow{8}{*}{GPT-4o Contextual} & ES & 34.53 & 88.64 & 85.38 & 47.86 & 89.45 & {\cellcolor[rgb]{0.812,0.812,0.812}}85.45 & {\cellcolor[rgb]{0.812,0.812,0.812}}50.96 & 91.40 & {\cellcolor[rgb]{0.812,0.812,0.812}}81.30 & 43.85 & {\cellcolor[rgb]{0.812,0.812,0.812}}87.96 & 77.34 & 42.77 & 85.23 & 80.08 \\
 & FR & 43.62 & 89.71 & {\cellcolor[rgb]{0.812,0.812,0.812}}85.81 & 43.70 & 85.17 & 80.60 & 50.10 & {\cellcolor[rgb]{0.812,0.812,0.812}}91.11 & 81.14 & {\cellcolor[rgb]{0.812,0.812,0.812}}47.76 & {\cellcolor[rgb]{0.812,0.812,0.812}}88.31 & 77.10 & 47.18 & 89.53 & 84.07 \\
 & IT & 38.48 & 89.11 & 85.43 & - & - & - & 50.33 & 91.46 & 80.87 & 46.63 & 88.93 & 76.68 & 46.93 & 89.82 & 84.00 \\
 & DE & 48.32 & 90.27 & {\cellcolor[rgb]{0.812,0.812,0.812}}85.78 & - & - & - & 50.27 & {\cellcolor[rgb]{0.812,0.812,0.812}}90.85 & {\cellcolor[rgb]{0.812,0.812,0.812}}81.26 & {\cellcolor[rgb]{0.812,0.812,0.812}}48.10 & 88.61 & 76.90 & 48.99 & {\cellcolor[rgb]{0.812,0.812,0.812}}90.02 & {\cellcolor[rgb]{0.812,0.812,0.812}}84.19 \\
 & RU & 46.65 & 89.97 & {\cellcolor[rgb]{0.812,0.812,0.812}}85.58 & 47.48 & 89.34 & 85.14 & 48.82 & 90.67 & 80.86 & 45.31 & 88.40 & 76.76 & 46.70 & 89.34 & 83.76 \\
 & ES + FR & 48.59 & {\cellcolor[rgb]{0.812,0.812,0.812}}90.50 & \textbf{86.03} & 51.73 & 90.04 & {\cellcolor[rgb]{0.812,0.812,0.812}}85.55 & 55.67 & 92.27 & 81.68 & {\cellcolor[rgb]{0.812,0.812,0.812}}48.40 & 88.76 & {\cellcolor[rgb]{0.812,0.812,0.812}}77.74 & {\cellcolor[rgb]{0.812,0.812,0.812}}50.65 & 90.47 & 84.52 \\
 & FR + IT & 48.85 & {\cellcolor[rgb]{0.812,0.812,0.812}}90.46 & {\cellcolor[rgb]{0.812,0.812,0.812}}85.88 & - & - & - & 54.64 & 92.08 & {\cellcolor[rgb]{0.812,0.812,0.812}}81.47 & 49.12 & 89.11 & 77.34 & {\cellcolor[rgb]{0.812,0.812,0.812}}50.40 & 90.39 & {\cellcolor[rgb]{0.812,0.812,0.812}}84.48 \\
 & ES + FR + IT & 49.14 & {\cellcolor[rgb]{0.812,0.812,0.812}}90.57 & 86.02 & - & - & - & \textbf{56.48} & \textbf{92.39} & 81.74 & \textbf{49.35} & \textbf{89.15} & {\cellcolor[rgb]{0.812,0.812,0.812}}77.67 & \textbf{51.46} & \textbf{90.64} & 84.52 \\ 
\hline
\multirow{8}{*}{GPT-4o Sequential} & ES & 48.81 & 90.23 & {\cellcolor[rgb]{0.812,0.812,0.812}}85.81 & 50.87 & 90.07 & {\cellcolor[rgb]{0.812,0.812,0.812}}85.55 & 50.25 & {\cellcolor[rgb]{0.812,0.812,0.812}}90.89 & 81.80 & 45.52 & 87.36 & {\cellcolor[rgb]{0.812,0.812,0.812}}77.67 & 48.76 & {\cellcolor[rgb]{0.812,0.812,0.812}}89.93 & 84.52 \\
 & FR & 48.02 & 90.16 & {\cellcolor[rgb]{0.812,0.812,0.812}}85.81 & 50.98 & 90.16 & {\cellcolor[rgb]{0.812,0.812,0.812}}\textbf{85.62} & {\cellcolor[rgb]{0.812,0.812,0.812}}50.90 & {\cellcolor[rgb]{0.812,0.812,0.812}}91.06 & 81.63 & 45.09 & 87.36 & {\cellcolor[rgb]{0.812,0.812,0.812}}77.70 & 48.18 & 89.77 & 84.53 \\
 & IT & 48.00 & 90.12 & {\cellcolor[rgb]{0.812,0.812,0.812}}85.70 & - & - & - & {\cellcolor[rgb]{0.812,0.812,0.812}}50.86 & {\cellcolor[rgb]{0.812,0.812,0.812}}91.12 & {\cellcolor[rgb]{0.812,0.812,0.812}}81.42 & 45.76 & 87.78 & 77.30 & 48.79 & {\cellcolor[rgb]{0.812,0.812,0.812}}89.89 & {\cellcolor[rgb]{0.812,0.812,0.812}}84.45 \\
 & DE & 49.72 & 90.34 & {\cellcolor[rgb]{0.812,0.812,0.812}}85.76 & - & - & - & {\cellcolor[rgb]{0.812,0.812,0.812}}50.93 & {\cellcolor[rgb]{0.812,0.812,0.812}}91.02 & {\cellcolor[rgb]{0.812,0.812,0.812}}81.57 & 46.32 & 87.85 & {\cellcolor[rgb]{0.812,0.812,0.812}}77.83 & 49.30 & {\cellcolor[rgb]{0.812,0.812,0.812}}90.03 & 84.51 \\
 & RU & 49.21 & 90.38 & {\cellcolor[rgb]{0.812,0.812,0.812}}85.81 & 50.64 & 90.06 & {\cellcolor[rgb]{0.812,0.812,0.812}}85.55 & {\cellcolor[rgb]{0.812,0.812,0.812}}51.22 & {\cellcolor[rgb]{0.812,0.812,0.812}}91.03 & {\cellcolor[rgb]{0.812,0.812,0.812}}81.56 & 44.83 & 87.61 & {\cellcolor[rgb]{0.812,0.812,0.812}}77.60 & 48.60 & {\cellcolor[rgb]{0.812,0.812,0.812}}89.96 & {\cellcolor[rgb]{0.812,0.812,0.812}}84.45 \\
 & ES + FR & 49.98 & {\cellcolor[rgb]{0.812,0.812,0.812}}90.44 & 85.90 & 52.34 & {\cellcolor[rgb]{0.812,0.812,0.812}}90.31 & {\cellcolor[rgb]{0.812,0.812,0.812}}\textbf{85.62} & {\cellcolor[rgb]{0.812,0.812,0.812}}51.84 & {\cellcolor[rgb]{0.812,0.812,0.812}}91.16 & \textbf{81.81} & 46.54 & 87.52 & \textbf{77.89} & {\cellcolor[rgb]{0.812,0.812,0.812}}50.42 & {\cellcolor[rgb]{0.812,0.812,0.812}}90.10 & \textbf{84.58} \\
 & FR + IT & 49.74 & 90.41 & {\cellcolor[rgb]{0.812,0.812,0.812}}85.80 & - & - & - & {\cellcolor[rgb]{0.812,0.812,0.812}}51.74 & 91.30 & 81.64 & 47.02 & 87.81 & 77.86 & 49.93 & {\cellcolor[rgb]{0.812,0.812,0.812}}90.08 & 84.55 \\
 & ES + FR + IT & 50.26 & {\cellcolor[rgb]{0.812,0.812,0.812}}90.51 & 85.92 & - & - & - & 52.46 & 91.35 & 81.75 & 47.08 & 87.72 & {\cellcolor[rgb]{0.812,0.812,0.812}}77.73 & {\cellcolor[rgb]{0.812,0.812,0.812}}50.48 & {\cellcolor[rgb]{0.812,0.812,0.812}}90.17 & \textbf{84.58} \\ 
\hline\hline
NMT Baseline & NONE & \textbf{53.02} & \textbf{90.52} & \textbf{85.91} & \textbf{51.79} & \textbf{90.07} & \textbf{85.62} & \textbf{58.25} & \textbf{91.77} & \textbf{81.89} & \textbf{49.31} & \textbf{88.20} & \textbf{78.17} & \textbf{54.00} & \textbf{90.44} & \textbf{84.60} \\ 
\hline
\multirow{5}{*}{NMT Shallow Fusion} & ES & 45.82 & 89.84 & 85.34 & 48.91 & 89.53 & {\cellcolor[rgb]{0.812,0.812,0.812}}85.52 & 56.29 & 91.41 & 81.41 & 47.10 & 87.69 & 78.00 & 51.44 & 90.02 & 84.22 \\
 & FR & 48.91 & 90.05 & 85.34 & 43.46 & 85.75 & 80.50 & 55.88 & 91.28 & 81.34 & 47.60 & 87.89 & 77.79 & 50.73 & 89.89 & 84.20 \\
 & IT & 47.00 & 89.87 & 85.46 & - & - & - & 55.68 & 91.32 & 81.24 & 47.41 & {\cellcolor[rgb]{0.812,0.812,0.812}}87.99 & 77.70 & 49.76 & 89.86 & 84.02 \\
 & DE & 48.00 & 90.10 & 85.53 & - & - & - & 55.26 & 90.98 & 81.48 & 46.27 & 87.70 & 77.72 & 49.79 & 89.95 & 84.19 \\
 & ES + FR + IT & 44.83 & 89.76 & 85.51 & - & - & - & 56.29 & {\cellcolor[rgb]{0.812,0.812,0.812}}91.60 & 81.57 & 47.46 & {\cellcolor[rgb]{0.812,0.812,0.812}}88.13 & 77.96 & 50.05 & 90.07 & 84.20
\end{tabular}%
}
\caption{Complete English-to-Portuguese translation results for GPT-4o and the NMT system across proprietary datasets, TICO-19, and FLORES+, evaluated using BLEU, COMET, and COMETKIWI metrics. Baselines are provided for both GPT-4o and the NMT system. \textbf{Bold} values indicate the highest scores for each system (GPT-4o and NMT) in their respective experiments across all datasets. Results in \colorbox[rgb]{0.812,0.812,0.812}{gray} show no significant difference from the corresponding system's baseline according to each metric and dataset (p-value < 0.05).}
\label{tab:full_res_en_pt}
\vspace{2 cm}
\resizebox{\textwidth}{!}{%
\begin{tabular}{l|l|lll|lll|lll|lll|lll}
 &  & \multicolumn{3}{c|}{\textbf{FLORES+}} & \multicolumn{3}{c|}{\textbf{TICO-19}} & \multicolumn{3}{c|}{\textbf{PROPRIETARY A}} & \multicolumn{3}{c|}{\textbf{PROPRIETARY B}} & \multicolumn{3}{c}{\textbf{PROPRIETARY C}} \\
Model/Experiment & CONTEXT & BLEU & COMET & COMETKIWI & BLEU & COMET & COMETKIWI & BLEU & COMET & COMETKIWI & BLEU & COMET & COMETKIWI & BLEU & COMET & COMETKIWI \\ 
\hline
GPT-4o Baseline & NONE & 26.48 & 87.60 & \textbf{83.64} & 29.54 & 87.20 & \textbf{83.21} & 34.44 & 88.28 & \textbf{78.42} & 27.88 & 83.86 & \textbf{74.99} & 28.78 & 86.62 & \textbf{82.12} \\ 
\hline
\multirow{2}{*}{GPT-4o Contextual} & ES & 27.85 & 87.99 & 79.69 & 44.38 & 89.14 & 80.39 & 47.74 & 90.97 & 75.71 & 40.71 & 87.42 & 71.00 & 41.57 & 89.14 & 79.14 \\
 & ES + FR & \textbf{37.62} & \textbf{89.45} & 81.80 & \textbf{44.76} & \textbf{89.48} & 81.36 & \textbf{50.11} & \textbf{91.31} & 76.44 & \textbf{44.29} & \textbf{88.26} & 71.79 & \textbf{43.76} & \textbf{89.61} & 80.08
\end{tabular}%
}
\caption{Complete Chinese-to-Portuguese translation results for GPT-4o across proprietary datasets, TICO-19, and FLORES+, evaluated using BLEU, COMET, and COMETKIWI metrics. \textbf{Bold} values represent the highest scores compared to the baseline. All results are significantly different from the baseline across each metric and dataset (p-value < 0.05).}
\label{tab:full_res_zh_pt}
\end{sidewaystable*}